%% file: main.tex
\pdfoutput=1
\documentclass[conference]{IEEEtran}
\IEEEoverridecommandlockouts
\usepackage{cite}
\usepackage{amsmath,amssymb,amsfonts}
\usepackage{algorithmic}
\usepackage{graphicx}
\usepackage{textcomp}
\usepackage[table,xcdraw]{xcolor}
\usepackage{float} 
\usepackage{xurl}
\usepackage{multicol}
\usepackage{setspace}
\usepackage{threeparttable}

\def\BibTeX{{\rm B\kern-.05em{\sc i\kern-.025em b}\kern-.08em
    T\kern-.1667em\lower.7ex\hbox{E}\kern-.125emX}}
\begin{document}

% \title{Obstacle Tower Challenge:\\ How To Cope Without Human Demonstration}
\title{Obstacle Tower Without Human Demonstrations: How Far a Deep Feed-Forward Network Goes with Reinforcement Learning}
% OTC without human demonstration: how far a deep feed-forward network goes with RL
% with Reinforcement Learning
% how far a deep feed-forward network can go without human demonstration 
% can manage

%\author{\IEEEauthorblockN{Marco Pleines}
%\IEEEauthorblockA{\textit{Faculty of Computer Science} %\\
%\textit{Technische Universität Dortmund}\\
%Dortmund, Germany \\
%marco.pleines@tu-dortmund.de}
%\and
%\IEEEauthorblockN{Jenia Jitsev}
%\IEEEauthorblockA{\textit{Juelich Supercomputing Center %(JSC)} \\
%\textit{Helmholtz Research Center Juelich}\\
%Juelich, Germany \\
%j.jitsev@fz-juelich.de}
%\and
%\IEEEauthorblockN{Mike Preuss}
%\IEEEauthorblockA{\textit{LIACS} \\
%\textit{Universiteit Leiden}\\
%Leiden/NL \\
%m.preuss@liacs.leidenuniv.nl}
%\and
%\IEEEauthorblockN{Frank Zimmer}
%\IEEEauthorblockA{\textit{Faculty of Communication and %Environment} \\
%\textit{Hochschule Rhein-Waal}\\
%Kamp-Lintfort, Germany \\
%frank.zimmer@hochschule-rhein-waal.de}
%}

\author{
\IEEEauthorblockN{Marco Pleines\IEEEauthorrefmark{1}
, Jenia Jitsev\IEEEauthorrefmark{2}, Mike Preuss
\IEEEauthorrefmark{3}, Frank Zimmer\IEEEauthorrefmark{4}}
\IEEEauthorblockA{
\IEEEauthorrefmark{1}\textit{Faculty of Computer Science},
\textit{Technische Universität Dortmund},
Dortmund, Germany \\
marco.pleines@tu-dortmund.de}
\IEEEauthorblockA{
\IEEEauthorrefmark{2}\textit{Juelich Supercomputing Center (JSC)},
\textit{Helmholtz Research Center Juelich},
Juelich, Germany \\
j.jitsev@fz-juelich.de}
\IEEEauthorblockA{
\IEEEauthorrefmark{3}\textit{LIACS},
\textit{Universiteit Leiden},
Leiden/NL \\
m.preuss@liacs.leidenuniv.nl}
\IEEEauthorblockA{\IEEEauthorrefmark{4}\textit{Faculty of Communication and Environment},
\textit{Hochschule Rhein-Waal},
Kamp-Lintfort, Germany\\
frank.zimmer@hochschule-rhein-waal.de}
}

\maketitle

\input{content/abstract.tex}
\input{content/introduction.tex}
\input{content/related_work.tex}

\input{content/approach.tex}
\input{content/analysis.tex}
\input{content/discussion.tex}
\input{content/conclusion.tex}

%\begin{figure}[htbp]
%\centerline{\includegraphics{fig1.png}}
%\caption{Example of a figure caption.}
%\label{fig}
%\end{figure}

% disabled for first version:
%\section*{Acknowledgment}

%%%%%%%%%%%%%%%%%%%%%%%%%%%%%%%%%%%%%%%%
%%%%%%                      Bibliography
%%%%%%%%%%%%%%%%%%%%%%%%%%%%%%%%%%%%%%%%

%\nocite{*}
\bibliographystyle{./bibliography/IEEEtran}
\bibliography{./bibliography/IEEEabrv.bib,./bibliography/bibliography.bib}
\end{document}

%% file: content/abstract.tex
%%%%%%%%%%%%%%%%%%%%%%%%%%%%%%%%%%%%%%%%
%%%%%%                          Abstract
%%%%%%%%%%%%%%%%%%%%%%%%%%%%%%%%%%%%%%%%

\begin{abstract}
The Obstacle Tower Challenge is the task to master a procedurally generated chain of levels that subsequently get harder to complete.
Whereas the most top performing entries of last year's competition used human demonstrations or reward shaping to learn how to cope with the challenge, we present an approach that performed competitively (placed 7th) but starts completely from scratch by means of Deep Reinforcement Learning with a relatively simple feed-forward deep network structure. 
We especially look at the generalization performance of the taken approach concerning different seeds and various visual themes that have become available after the competition, and investigate where the agent fails and why.
Note that our approach does not possess a short-term memory like employing recurrent hidden states.
With this work, we hope to contribute to a better understanding of what is possible with a relatively simple, flexible solution that can be applied to learning in environments featuring complex 3D visual input where the abstract task structure itself is still fairly simple.
\end{abstract}


%% file: content/introduction.tex
\IEEEpubid{\begin{minipage}{\textwidth}\ \\ \\[12pt] 978-1-7281-4533-4/20/\$31.00 \copyright 2020 IEEE \end{minipage}}

%%%%%%%%%%%%%%%%%%%%%%%%%%%%%%%%%%%%%%%%
%%%%%%                      Introduction
%%%%%%%%%%%%%%%%%%%%%%%%%%%%%%%%%%%%%%%%
\section{Introduction}
\newcommand{\RNum}[1]{\uppercase\expandafter{\romannumeral #1\relax}}

% current state of reinforcement learning to narrow down the context of this paper
Deep Reinforcement Learning (DRL) had tremendous successes during the last years.
Very often it has been employed as direct end-to-end learning from high-dimensional raw pixel images for difficult tasks such as playing Atari games~\cite{Mnih2015}, Doom~\cite{Wydmuch2019}, or the more cooperative games capture-the-flag~\cite{Jaderberg2019} and Dota~2 ~\cite{OpenAI2019}.
Such game environments are also more and more combined with additional information beyond pure pixels as for hide-and-seek \cite{Baker2019} and AlphaStar~\cite{Vinyals2019}.
The latter one plays the complex real-time strategy game StarCraft\,\RNum{2} on human grandmaster level, a milestone that has been presumed not reachable for years not long ago.

Concerning Atari games, feed-forward convolutional neural networks (FFCNN) can be successfully trained to solve those using basic policy gradient methods as REINFORCE or value based ones as Q-Learning.
For more complex environments featuring 3D worlds or sparse long horizon rewards, more advanced network architectures are considered, like various convolutional recurrent neural networks.
This paper shows that up to a certain degree it is also possible to solve complex 3D environments using a rather simple FFCNN when training those with state-of-the art DRL algorithms like Proximal Policy Optimization (PPO)~\cite{Schulman2017}.

Furthermore, evaluating a model's generalization capability, obtained through DRL, usually lacks in a clear split between training and testing phases due to the widely used standard benchmark game environments that suffer from fixed structures.
To overcome this issue, environments that utilize procedural content generation (PCG) approaches shall be employed here.
Hence, the model can be trained and evaluated on distinct seeds, each defining a unique instance of an environment, guaranteeing a clear split.

One example for a procedurally generated environment is Obstacle Tower (OT)~\cite{Juliani2019}.
In OT, the agent is challenged in terms of vision, control, planning, and generalization, while its goal is to ascend a tower of floors that get more difficult as the agent progresses~\cite{Juliani2019}.
The first 5 floors do not involve any special puzzles for the agent.
After that, the agent has to find keys to get past locked doors.
Once floor 10 is reached, a difficult sokoban puzzle is introduced.
In 2019, the developers of OT held a challenge where the top entries moved beyond floor 10 only with the help of domain knowledge such as adding human demonstrations to the training data ~\cite{Juliani2019_blog}.

In this work, we demonstrate that OT can be solved up to floor 10 using a rather simple FFCNN when trained with advanced DRL techniques (PPO) without the use of human demonstrations. In the original paper, which introduced OT, the highest floor reached using a FFCNN and PPO was 5.
Reaching Level 10 is quite challenging given the rather complex OT 3D world environment and tasks like key-door puzzles or double jumps introduced from level 5 on.
Overall, our FFCNN-PPO algorithm performed competitive in the official OT challenge, ranking 7th\footnote{https://youtu.be/P2rBDHBHxcM}.
\begin{figure*}
	\centering
	\includegraphics[width=1.0\textwidth]{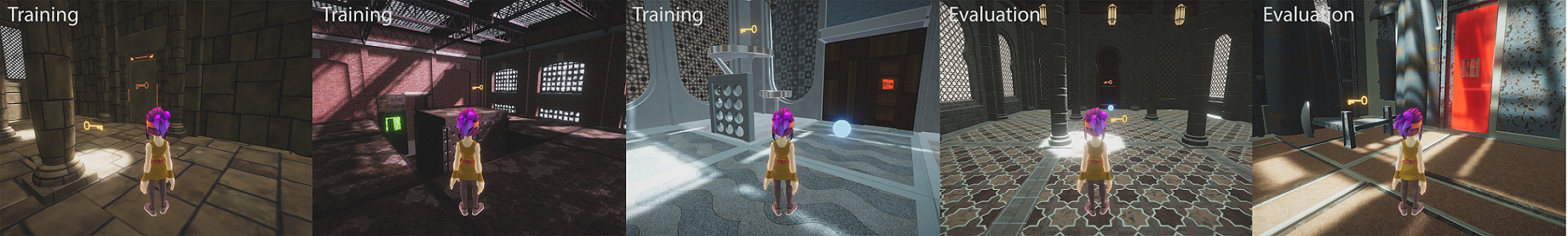}
	\caption{The 5 visual themes featured by Obstacle Tower from left to right: Ancient, Industrial, Modern, Moorish, and Future.}
	\label{img:visual_themes}
\end{figure*}

To further study generalization of our learning algorithm, we train our model on 3 and evaluate on 2 different visual themes (or skins) that are offered by the OT environment (Figure~\ref{img:visual_themes}).
Training on selected skin sets and testing on ones the algorithm has not seen before allows us to draw conclusions about its generalization capability with regard to the environment's vision challenge.
While we can state that the FFCNN is able to cope well with different visual themes during training without collapsing, we observe a clear drop in testing performance on the novel themes, which shows obvious generalization limits of the FFCNN.

This  paper  proceeds  as  follows: the next section highlights the measures taken by the top competitors of the OT challenge, while showcasing other work in the broader context of generalization.
Section 2 describes the taken approach concerning its details about the environment configuration, the architecture of the FFCNN, and the PPO training.
After that, the conducted experiment is described.
Results are shown for the generalization performance of the trained model as well as a detailed examination of the learned policy.
To further elaborate our achievements, section 5 discusses the observed peculiarities of our results and approach.
The last section concludes our findings and describes consecutive work.

%% file: content/related_work.tex
\section{Related Work}

We start by relating our work to the top competitors of the OT challenge before putting it into the wider context of generalization.
The challenge's organizers state that average human performance is around floor 15~\cite{Juliani2019}, and this has only been surpassed by the winner (average floor 19.4) and the runner-up (average floor 16) of the challenge.  
The top 4 entries were able to get past floor 10 by means of a PPO approach that was augmented with human demonstrations~\cite{Juliani2019_blog}. %\footnote{\url{https://blogs.unity3d.com/2019/08/07/announcing-the-obstacle-tower-challenge-winners-and-open-source-release/}}
Common measures, shared by several approaches, to reduce the problem complexity are:
\begin{itemize}
    \item Reduced action space: OT features a multi-discrete action space containing 3 subspaces comprising 11 actions in total. If a regular single discrete space is used, 54 possible action combinations become available. Many of those may not be necessary for achieving good results on the one hand. On the other hand, large action numbers also make the learning problem more difficult. It becomes conceivably easier to learn with only around 10 usable actions, by also preselecting reasonable action combinations of those that are potentially possible.
    \item Use of memory cells: simple neural networks, like the one employed in this work, have no means of treating developments over time in a meaningful way. However, this is possible with GRU or LSTM cells which are used by several related approaches.
    \item Frame stacks: adding past frames to the agent's observation is another popular measure, especially if no memory cell is used.
    \item Data augmentation: such techniques improve data quality and simplify the learning process. Mirroring left to right and vice versa is one example.
    \item Reward shaping: the learning process can be sped up by providing more and better suited rewards to the agent.
\end{itemize}

As of now, the 1st and 4th place shared technical details via blog posts and a preprint~\cite{Nichol2019, Booth2019}.
The competition winner, Alex Nichol, trained a classifier to help the agent detect various environment entities such as doors and keys.
Throughout 50 consecutive frames, Nichol adds the received reward, the executed action, the possession of the key, and the classifier's output to the agent's observation space.
Instead of using an entropy bonus for exploration, he applies KL-Divergence to push the agent's policy towards a prior, which was trained with behavioral cloning beforehand.
This way the agent was able to solve the complex sokoban puzzle.

%There is currently not much material available on the technical details of the top ranked architectures.
%However, the competition winner, Alex Nichol (unixpickle) has published a blog post\footnote{\url{ https://blog.aqnichol.com/2019/07/24/competing-in-the-obstacle-tower-challenge/}} and his github repository\footnote{\url{https://github.com/unixpickle/obs-tower2}}, providing a lot of detailed information about his entry.
%Joe Booth, placed 4th, has published a blog post\footnote{\url{https://towardsdatascience.com/i-placed-4th-in-my-first-ai-competition-takeaways-from-the-unity-obstacle-tower-competition-794d3e6d3310}} and also a preprint with technical details\footnote{\url{https://arxiv.org/abs/1907.06704}}.

One of the main motivations for setting up the OT challenge was to see how learning methods can deal with generalization.
By exposing learning agents to highly variable environments, overfitting~\cite{Bishop2006, zhang2018} shall be reduced and the agents should focus on learning the underlying major factors and not specific details of a single problem instance that are often misguiding in general cases. 
Encouraging generalization of the learning process by injecting more diversity into training environments has also been the main motivation to set up the GVGAI~\cite{perez2019general} environment.
There, the diverse set of games and levels has been created manually at first.
However, the setup also blends well with procedural content generation (PCG) techniques \cite{pcgbook} which are designed to provide controllable content variations that can be introduced systematically and automatized. 

Several works have investigated how PCG can be used in order to strengthen generalization~\cite{Justesen2018, Cobbe2019a}. 
Learning environments, like Procgen~\cite{Cobbe2019}, explicitly focus on enabling this and offer benchmarks for testing the generalization ability of RL algorithms.
Other approaches for achieving stronger generalization consist of, for example, adding  different types of memory to the neural networks~\cite{FortunatoNIPS2019}, inject noise~\cite{HofmannNIPS2019}, randomize the network's feature space ~\cite{lee2019network}, or randomize and distort the raw visual input from the training domain~\cite{James2019}.

% Generalization in other PCG domains Procgen/CoinRun

% other generalization studies
    % http://papers.nips.cc/paper/9546-generalization-in-reinforcement-learning-with-selective-noise-injection-and-information-bottleneck

% Other environments in terms of PCG and generalization
    % Procgen / CoinRun
    % The Animal-AI Environment: Training and Testing Animal-Like Artificial Cognition
    % Video Game Description Language Environment
    % Rogoue gym
        % https://ieeexplore.ieee.org/abstract/document/8848075/
    % Gym-Minigrid

%% file: content/approach.tex
%%%%%%%%%%%%%%%%%%%%%%%%%%%%%%%%%%%%%%%%
%%%%%%                          Approach
%%%%%%%%%%%%%%%%%%%%%%%%%%%%%%%%%%%%%%%%
\section{Approach}

% Limitations and constraints
    % Obstacle Tower slow simluation speed
        % Run 100 episodes, count steps, divide by seconds
        % Procgen Coinrun: 5375 steps/second
        % Minigrid: 634 steps/second
        % OT: 43 steps/second (Floor generation 100),  (Floor generation 10)
        % Hardware: NVIDIA Quadro K1200 4GB, 2x Intel Xeon E5-2640v4 (10x 2,4GHz), 64GB RAM
    % Long training durations

\begin{table}[]
    \begin{small}
    \centering
    \begin{tabular}{l|r}
        \rowcolor[HTML]{D1D1D1}
        \textbf{Environment}&\textbf{Steps per Second}  \\ \hline
        \rowcolor[HTML]{EFEFEF}
        Procgen CoinRun     & 5375                      \\ 
        \rowcolor[HTML]{D1D1D1}
        MiniGrid FourRoom \cite{gym_minigrid}  & 634                       \\ 
        \rowcolor[HTML]{EFEFEF}
        Atari Breakout \cite{Bellemare2013}     & 4041                      \\ 
        \rowcolor[HTML]{EFEFEF}
        Obstacle Tower (100 floors)      & 43           \\ 
        \rowcolor[HTML]{D1D1D1}
        Obstacle Tower (10 floors)   & 51               \\ 
    \end{tabular}
    \caption{The number of steps per second is averaged over 100 episodes per environment.}
    \label{tab:simulation_speed}
    \end{small}
\end{table}

Before elaborating the taken approach in greater detail, it is important to show that the simulation speed of the OT environment limits the number of training sessions and experiments.
As seen in Table~\ref{tab:simulation_speed}, OT runs much slower than other environments.
The benchmarks were run on an Ubuntu machine (nVidia Quadro K1200, 2x Intel Xeon E5-2640v4 CPUs, 64GB RAM).
In order to speed up the training process, the floor generation is limited to 10 floors.
This means that once the agent completes floor 9, the episode terminates with the result, that the agent reached floor 10.
Therefore, the difficult sokoban puzzle is not part of the training.

%%%%%%%%%%%%%%%%%%%%%%%%%%%%%%%%%%%%%%%%
\subsection{Environment Properties}

% Environment
    % frame skipping = 2
    % frame stacks = 3
    % visual and vector inputs
    % reduced action space
        % no-op, move forward
        % no-op, rotate left, rotate right
        % no-op, jump
    % Rewards
        % +1 exit floor
        % +0.1 collect key
        % +0.1 open door
    % 3 level designs for training
    % 2 level designs for evaluation
    % 100 seeds for training
    % 5 seeds for evaluation
    % each seed is evaluated 3 times
    % otherwise default OTC settings

Besides limiting the number of generated floors, we apply further changes to the environment.
To simplify the challenge, the agent executes one action for two consecutive frames (i.e. frame skipping).
Its action space is reduced from 11 to 7 actions, which are represented by 3 subspaces:
\begin{itemize}
    \begin{multicols}{2}
    \item Subspace A:
    \begin{itemize}
        \item No action
        \item Move forward
    \end{itemize}
    \item Subspace B:
    \begin{itemize}
        \item No action
        \item Jump
    \end{itemize}
    \bigskip
    \item Subspace C:
    \begin{itemize}
        \item No action
        \item Rotate left
        \item Rotate right
    \end{itemize}
    \end{multicols}
\end{itemize}
Moving left, right and backward are removed from the original action space, because these are not mandatory to solve OT.
At last, the reward function remains unchanged:
\begin{itemize}
    \item $+1.0$ for reaching the next floor,
    \item $+0.1$ for opening a door,
    \item and $+0.1$ for collecting a key.
\end{itemize}
Concerning the observation space, the agent receives the current and the past two visual observations of the environment.
The stacked image frames shall enable the agent to derive its velocity and acceleration.
Turning the frames into gray-scale is not considered, because it might raise the difficulty for the agent to identify a key.
However, the RGB image frames were unintentionally divided by 255 twice instead of just once.
Finally, the agent receives a vector of game state variables, featuring the remaining time and whether the agent has a key or not.

%%%%%%%%%%%%%%%%%%%%%%%%%%%%%%%%%%%%%%%%
\subsection{Model Architecture}

% Network Structure
    % CNN
    % concatenation of flattened features and vector input
    % action branching (multiple policy heads)
    % shared parameters for value and policy heads
    % parameters: 2,134,592 (conv layers 3648)

\begin{figure*}
	\centering
	\includegraphics[width=0.75\textwidth]{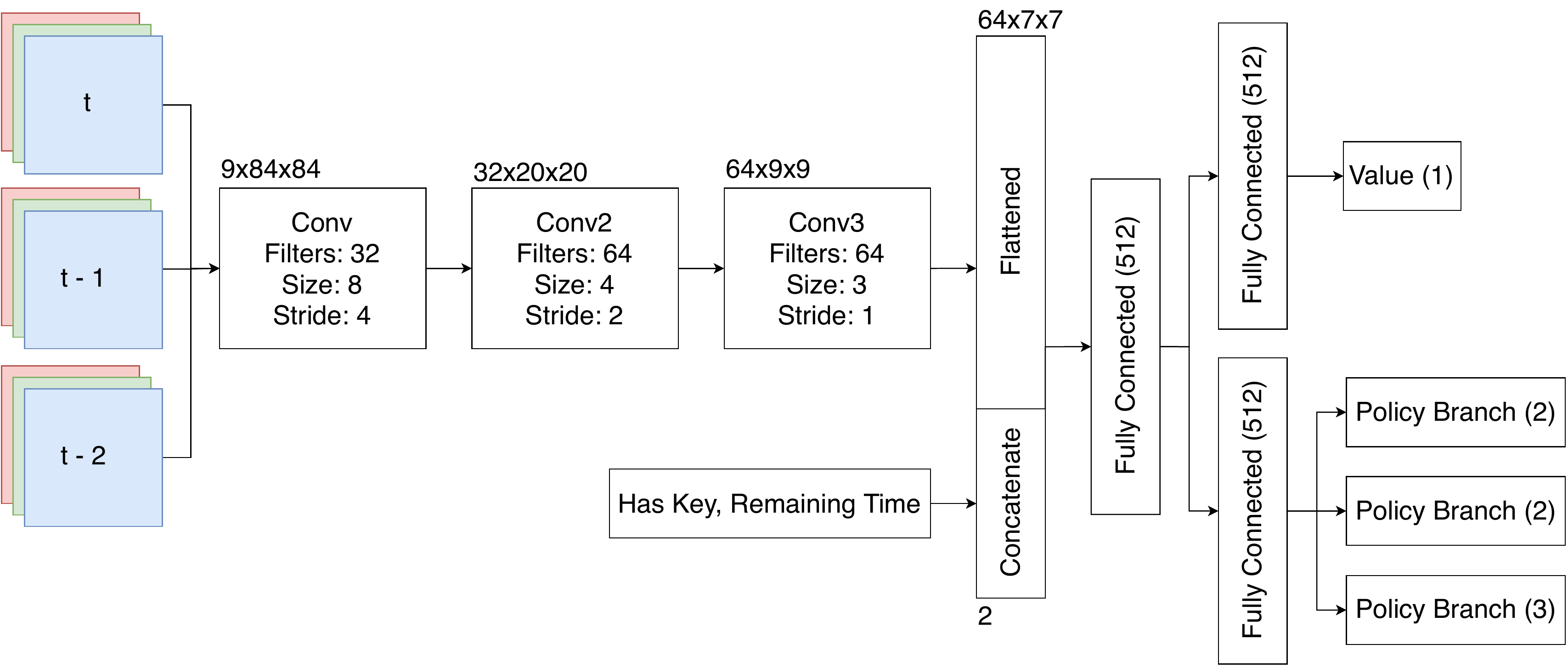}
	\caption[Model]{The architecture of the utilized feed-forward convolutional neural network.}
	\label{fig:model}
\end{figure*}

The environment properties affect the architecture of the trained FFCNN, which is illustrated by Figure~\ref{fig:model}.
The model receives few (n=3) temporally stacked image frames (each with 3 RGB color channels) and a vector of game state variables (has key, remaining time) as input.
Once the visual observation is processed by the convolutional layers (i.e. visual encoder), the flattened results and the game state vector input are concatenated.
The concatenation is then fed into a fully connected hidden layer, after which the neural net is split into two branches.
Those follow then the actor-critic architecture design~\cite{Sutton2018, Mnih2016}.
One branch is used for the value function that predicts the expected long-term reward with a single scalar output and the another one represents the policy signaling action probabilities.
Each branch contains a fully connected hidden layer.
Due to this setup, the value function and the policy share parameters of a common hidden layer while also maintaining a separate one in their respective branch.
As the action space is decomposed into 3 subspaces, the policy is composed of 3 branches as inspired by action branching~ \cite{Tavakoli2017}.
Each action branch in turn contains different numbers of available actions that predefine generic reasonable combinations (see Figure~\ref{fig:model}).
Hence, the model supports multi-discrete action spaces and avoids the necessity of implementing all action combinations causing a higher dimensional output.

%%%%%%%%%%%%%%%%%%%%%%%%%%%%%%%%%%%%%%%%
\subsection{PPO Training}

% Training
    % PPO (minibatch updates of trajectories at fixed length)
        % maybe show algorithm as listing, otherwise show only objectives and refer to the paper
    % clipped objectives
    % entropy bonus
    % how action branching affects the policy loss and the entropy bonus...
    % Generalized Advantage Estimation
    % Linearly annealing learning rate
    % Hyperparameters

\begin{table}[]
\begin{small}
\centering
    \begin{tabular}{l|r}
        \rowcolor[HTML]{D1D1D1} 
        {\color[HTML]{000000} \textbf{Training Parameter}}           & {\color[HTML]{000000} \textbf{Value}} \\ \hline
        \rowcolor[HTML]{EFEFEF} 
        {\color[HTML]{000000} Discount Factor}           & {\color[HTML]{000000} 0.99} \\
        \rowcolor[HTML]{D1D1D1} 
        Lamda (GAE)                                      & 0.95                        \\
        \rowcolor[HTML]{EFEFEF} 
        Value Function Coefficient                       & 0.5                         \\
        \rowcolor[HTML]{D1D1D1} 
        {\color[HTML]{000000} Entropy Bonus Coefficient} & {\color[HTML]{000000} 0.01} \\
        \rowcolor[HTML]{EFEFEF} 
        PPO Updates                                      & 50,000                      \\
        \rowcolor[HTML]{D1D1D1} 
        Epochs                                           & 4                           \\
        \rowcolor[HTML]{EFEFEF} 
        Number of Environments                           & 16                          \\
        \rowcolor[HTML]{D1D1D1} 
        Trajectory Length                                & 8,192                       \\
        \rowcolor[HTML]{EFEFEF} 
        Minibatches                                      & 4                           \\
        \rowcolor[HTML]{D1D1D1} 
        Learning Rate                                    & 3.25e-4                     \\
        \rowcolor[HTML]{EFEFEF} 
        Clip Range                                       & 0.2                         \\
        \rowcolor[HTML]{D1D1D1} 
        Activations                                      & ReLU                         \\
        \rowcolor[HTML]{EFEFEF}
        Optimizer                                        & Adam
    \end{tabular}
    \caption{Training Parameters}
    \label{tab:parameters}
\end{small}
\end{table}

The implementation\footnote{\url{https://github.com/MarcoMeter/neroRL}} of the training algorithm PPO is closely related to its publication from Schulman et al. (2017).
Generalized advantage estimation (GAE) is used by the value loss function.
The objectives for the value function and the policy are clipped.
Further, the final loss function comprises an entropy bonus term to encourage exploration.
However, action branching requires one small adjustment to the policy and the entropy bonus.
For each policy branch, all outputs are concatenated leading to a flattened view of these.
These are then processed by the loss function without further adjustments needed.
Concerning the entropy bonus, the mean of the policy branches' entropies is made use of.

Because of the high computational cost of running the OT environment, training parameters cannot undergo further optimization.
The ones provided by Table~\ref{tab:parameters} are derived from the experiences made during the OT challenge.
It has to be noted that the learning rate, the entropy bonus coefficient, and the clip range decay linearly dependent on the remaining PPO updates.
One PPO update optimizes the model using 4 minibatches per epoch across the training data, which was collected by the agents sampling actions from the current policy.
The intention behind annealing training parameters is to boost the agent's training performance in the beginning and then later to take smaller steps to fine-tune its policy.

%% file: content/analysis.tex
%%%%%%%%%%%%%%%%%%%%%%%%%%%%%%%%%%%%%%%%
%%%%%%             Experimental Analysis
%%%%%%%%%%%%%%%%%%%%%%%%%%%%%%%%%%%%%%%%
\section{Experimental Analysis}

Under the standard conditions of the OT environment, every 10th floor enables another visual theme to confront the agent.
So starting from floor 0 the agent only faces the ancient theme.
Once floor 10 is reached, the ancient and the moorish themes are alternated randomly.
Throughout the OT competition, 100 training seeds were available, while 5 distinct seeds were kept hidden to evaluate the agent's ability to generalize.
Therefore, the agents faced only the ancient and moorish theme during the competition's evaluation.
In order to fully assess the agent's generalization capability, we utilize three skins (ancient, industrial, and modern) for training, while leaving out the other ones (moorish, and future) for evaluation.
Due to the simulation speed constraints, we only train on this set as denoted by Figure~\ref{img:visual_themes}.

%%%%%%%%%%%%%%%%%%%%%%%%%%%%%%%%%%%%%%%%
\subsection{Generalization Performance}

\begin{figure}
	\centering
	\includegraphics[width=0.5\textwidth]{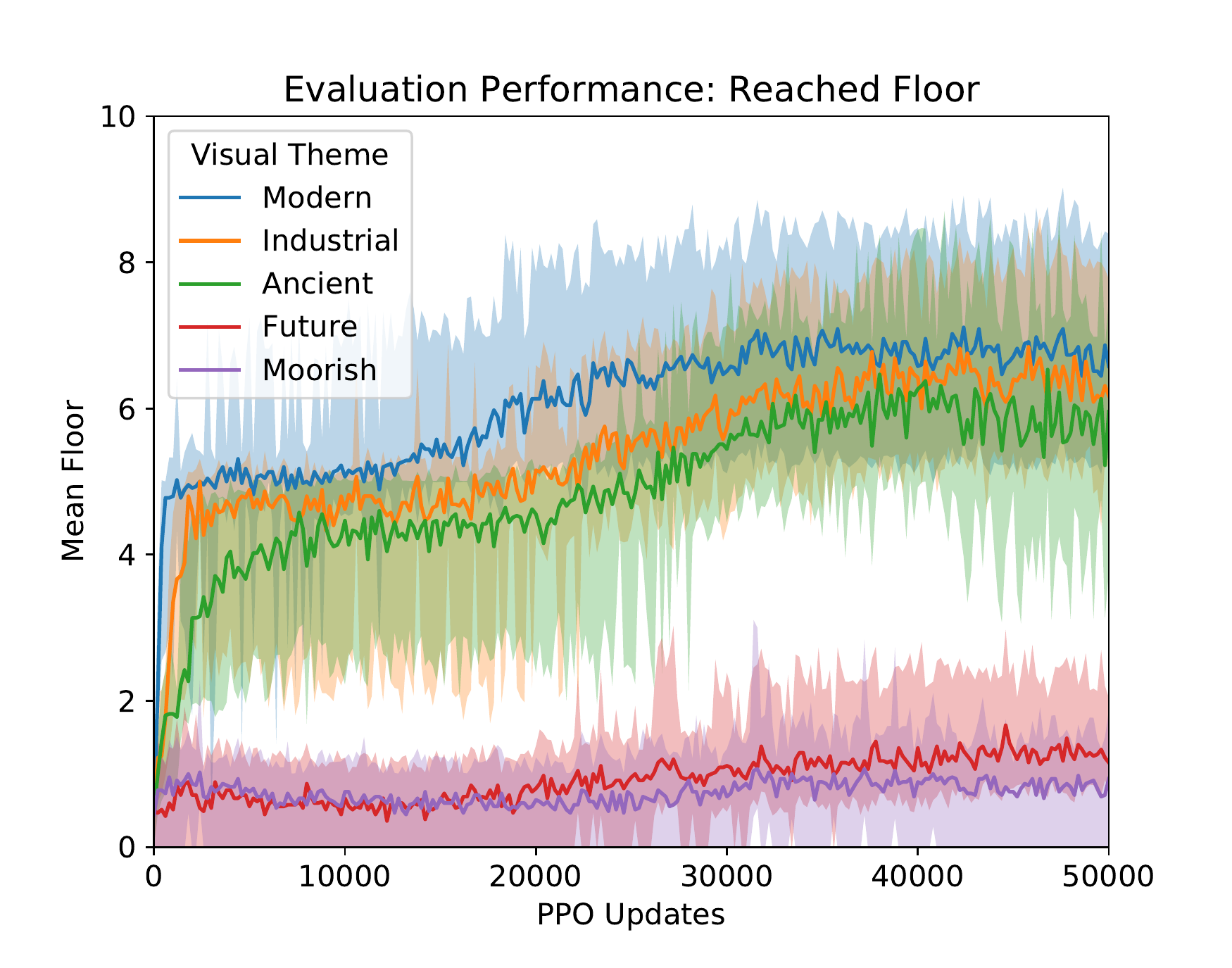}
	\caption{The achieved mean floor on all 5 visual themes across three training runs. Less opaque colored regions visualize their respective variance based on the asymmetric deviation. The moorish and the future theme were not part of the training.}
	\label{img:mean_floor}
\end{figure}

\begin{figure}
	\centering
	\includegraphics[width=0.42\textwidth]{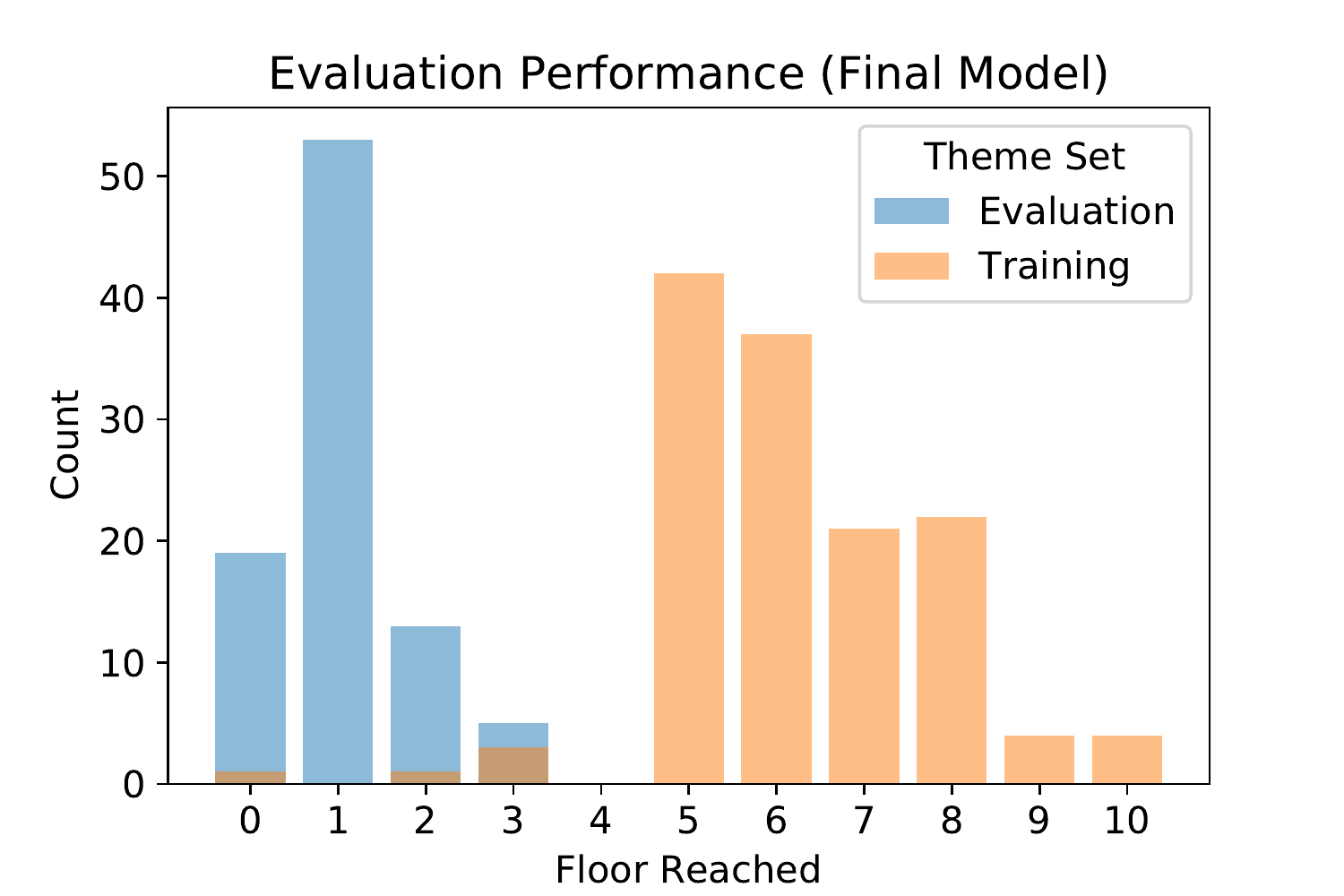}
	\caption{This bar chart visualizes the number of times an episode terminated at a certain floor on each theme set given the final model after training. A much lower performance is observed for the evaluation theme set.}
	\label{img:bar_chart}
\end{figure}

\begin{figure}
	\centering
	\includegraphics[width=0.5\textwidth]{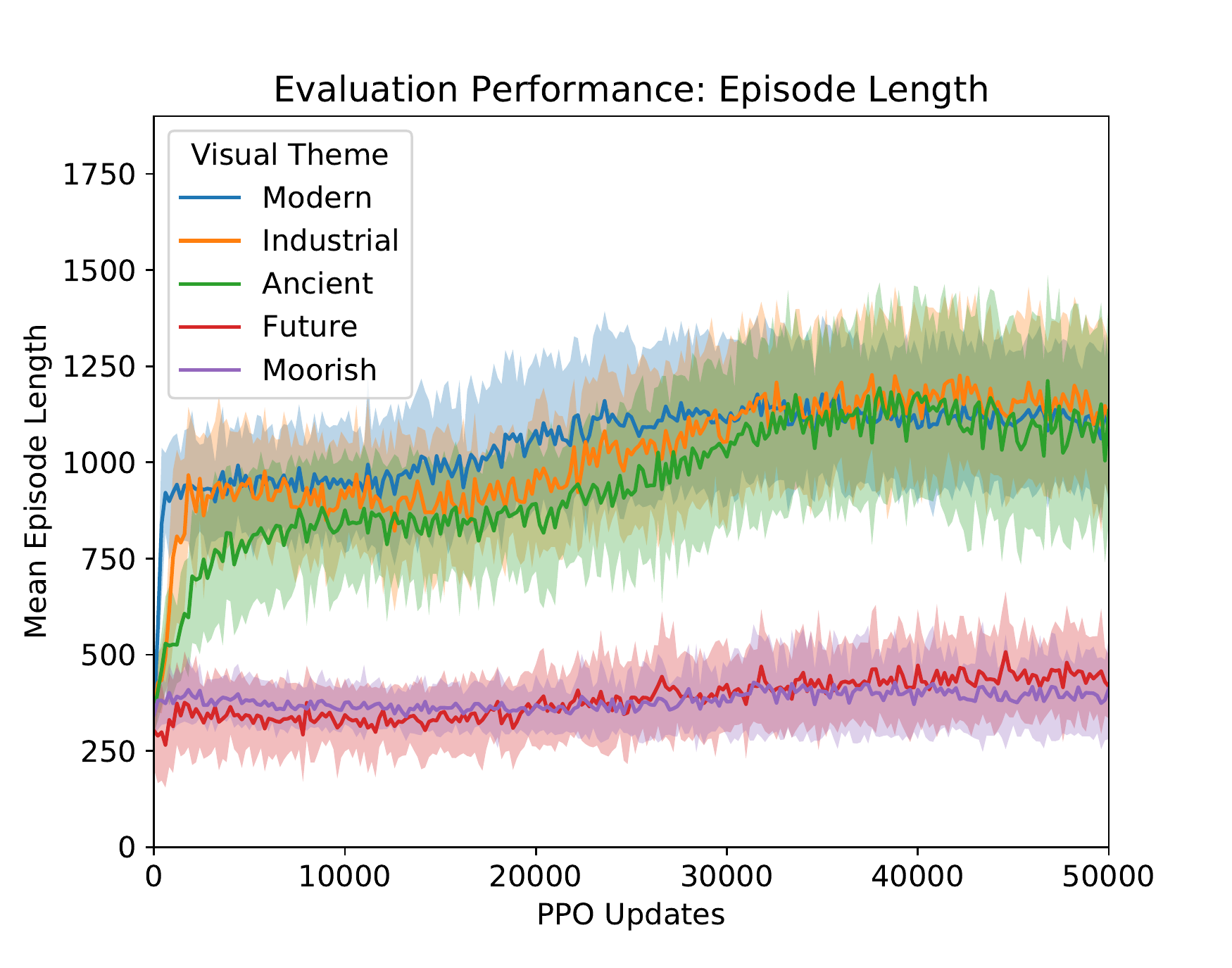}
	\caption{The achieved mean episode length on all 5 visual themes across three training runs. Less opaque colored regions visualize their respective variance based on the standard deviation. The moorish and the future theme were not part of the training.}
	\label{img:mean_length}
\end{figure}

We ran three training sessions using the same training parameters where the agent faced 100 seeds, while all three training skins were randomly alternated.
%It has to be noted that a change of skin affects the entire floor generation.
%Therefore, it can be derived that we trained ultimately on 300 seeds.
The only other modification to the environment is the limited floor number of 9.
While training for 50,000 PPO updates, every 200th update, the model was evaluated on 5 distinct seeds, which were not used during training.
These seeds were evaluated 3 times for all 5 visual themes.
Thus, the agent was evaluated by 15 episodes per theme.
Multiplied by three training runs, the results of 75 episodes were collected for each interval.
Figure~\ref{img:mean_floor} shows the achieved mean floor for each particular skin, whereas Figure~\ref{img:mean_length} illustrates the mean episode length.

It can be observed that the agent poorly performs on the moorish and the industrial theme, which were not seen by the agent during training.
By the end of the training, the agent reaches about mean floor 1.15 and 0.93 on these themes.
A different picture emerges on the training themes that the agent has seen during training.
On these, the agent's performance converges at about mean floor 6, although it did not encounter these seeds during training.
On the training seeds, the agent achieved a mean floor of about 8.66.
Overall, a high variance can be observed.
For instance on the ancient theme given the final model after training, the agent got stuck at floor 0 on one of the 15 evaluation episodes.
This variance becomes clearer by examining the number of times an episode terminated on a certain floor.
In the provided bar chart (Figure~\ref{img:bar_chart}), most of the time the episode ends on floor 5 on the training themes.
Terminations on floor 0, 2, 3, 9, and 10 can be understood as outliers.
Thus, there is always a chance that the agent accomplishes all 9 floors or gets stuck already at the first one, even if the same seed is tried over and over again.

During the first 10,000 PPO updates, the agent rapidly learns to reach floor 5 on the training skins.
Due to the introduction of the key puzzle tasks, the agent's policy is stuck for approximately 10,000 PPO updates on a plateau.
After that, the policy slowly improves over time.

Concerning the episode length, it is correlated with the successful tower ascend of the agent.
Reaching a new floor is rewarded with a time extension.
As the episode ends once floor 9 is finished, the episode length should decrease as the policy improves.
A slight trend for such decrease can be observed on the training skins, indicating that the agent becomes more proficient during its tower ascend.

%%%%%%%%%%%%%%%%%%%%%%%%%%%%%%%%%%%%%%%%
\subsection{Agent Behavior}

Multiple observations can be done by watching how the agent utilizes its learned policy to solve OT on the training themes.
First of all, it can be noticed that the agent's locomotion is rather shaky.
While moving forward, the agent tends to continuously execute the actions "rotate left" and "rotate right".
Another observation is that the agent likes to go through doors in general, no matter whether required or not for the current task. Therefore, it may also happen that the agent moves all the way back to the beginning instead of heading to the floor exit, but still the agent might be able to finish the current floor.
Regarding the jump action, the agent usually jumps when required, like in situations where the agent approaches an obstacle.
On very rare occasions, the agent gets stuck on corners of a door or on similar environmental structures while experiencing a rewarding stimulus in its visual field.
In this situation, the agent keeps uselessly moving forward and is therefore not able to finish the current floor.

\begin{figure}
	\centering
	\includegraphics[width=0.5\textwidth]{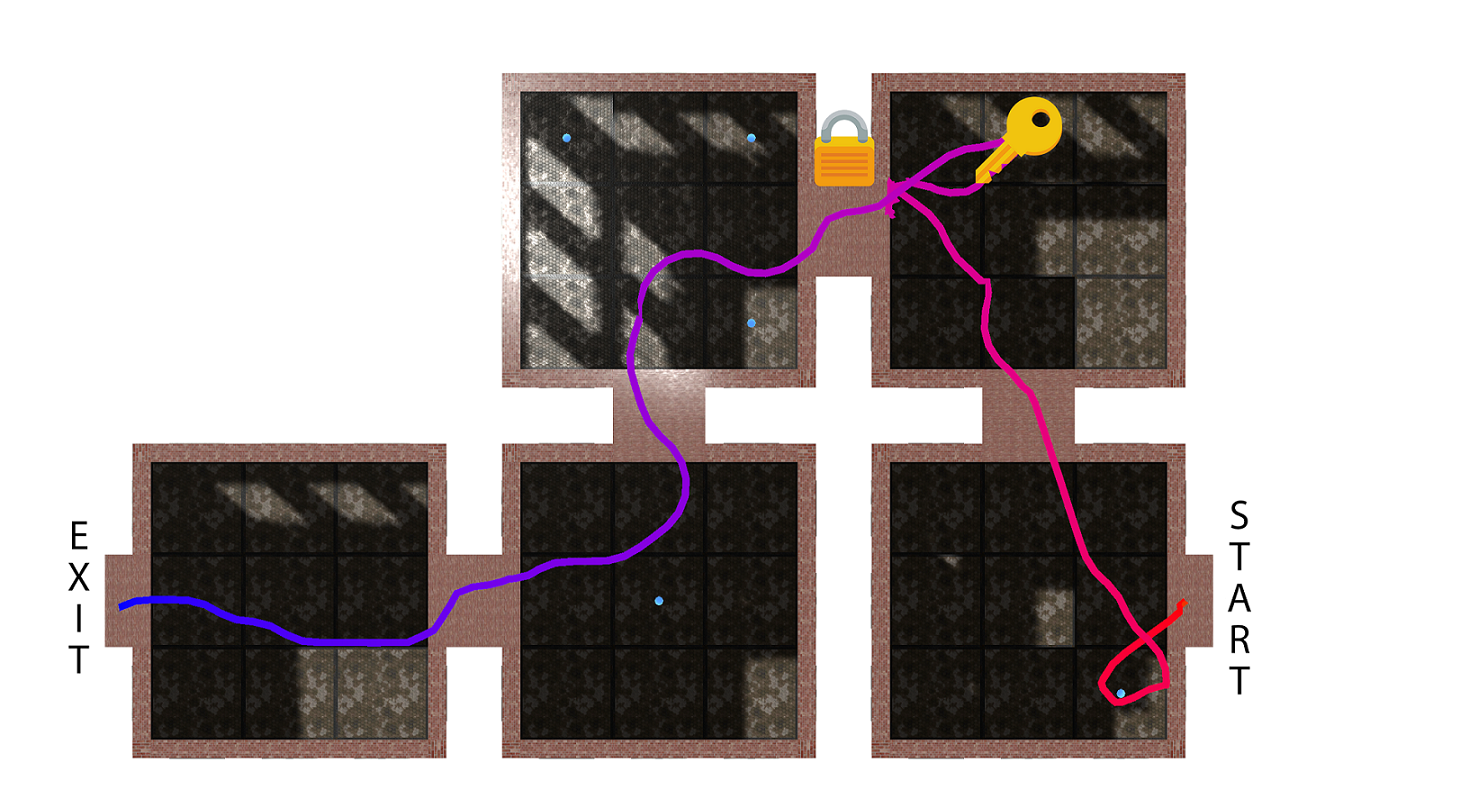}
	\caption{The agent immediately solves the key puzzle and exits the floor.}
	\label{img:path_A}
\end{figure}

\begin{figure}
	\centering
	\includegraphics[width=0.5\textwidth]{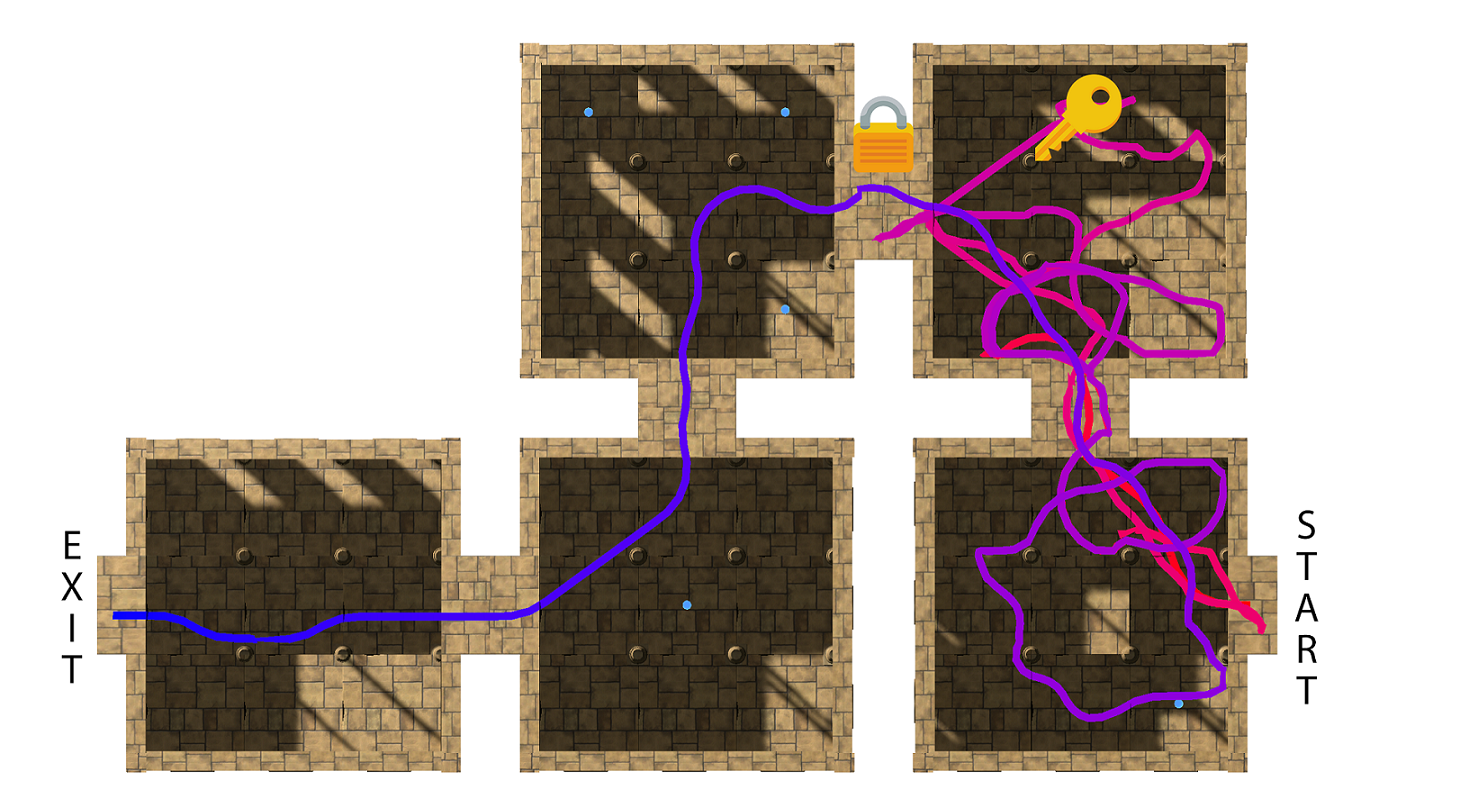}
	\caption{The agent wanders around the entire accessible rooms, but eventually collects the key and exits the floor.}
	\label{img:path_B}
\end{figure}

\begin{figure}
	\centering
	\includegraphics[width=0.475\textwidth]{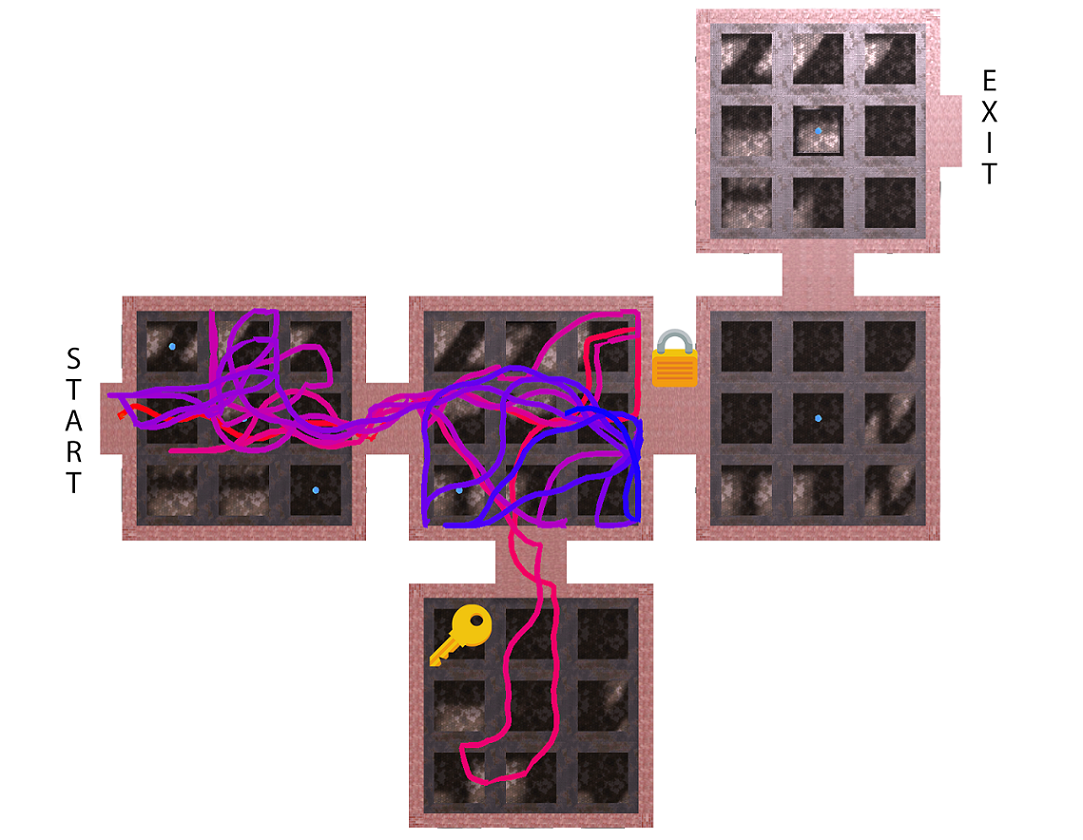}
	\caption{The agent wanders around the entire first two rooms. The key room is visited once, but the key was not collected. Thus, the floor was not solved.}
	\label{img:path_C}
\end{figure}

\begin{figure}
	\centering
	\includegraphics[width=0.4\textwidth]{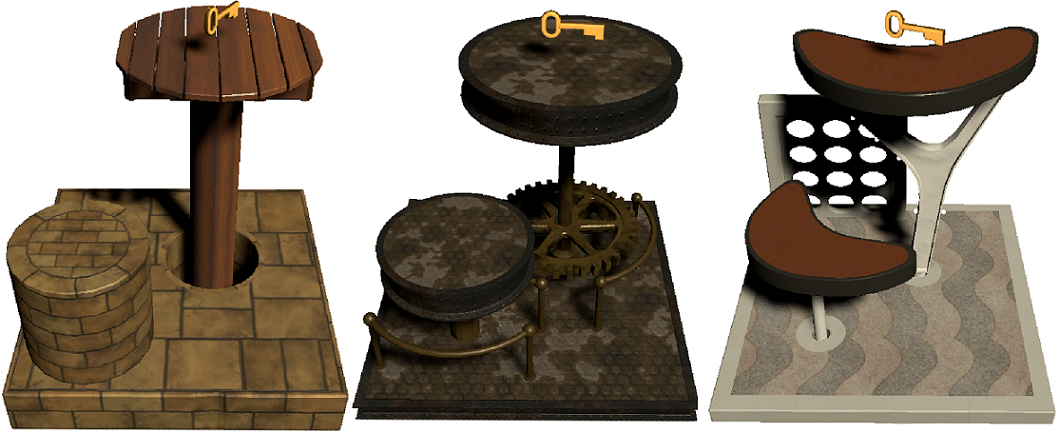}
	\caption{These are double jump modules used by the ancient, industrial, and modern theme, which were explored by the agent during training.}
	\label{img:double_jump}
\end{figure}

An important subtask, tackled by the agent, is the key puzzle, which can be partially visualized by the agent's taken path as seen in Figures \ref{img:path_A}, \ref{img:path_B}, and \ref{img:path_C}.
The drawn path of the agent starts out red and turns blue over time.
Further, the positions of the key and the door are marked by their respective icons.
Watching the agent's behavior, the paths shown by Figure \ref{img:path_B}, and \ref{img:path_C} are more likely to occur.
Most of the time the agent walks across the entire accessible space to grab the key by chance.
Also, it can be observed that the agent tries to get past locked doors without being in possession of the key.
It may happen that the agent attempts to pass the same locked door numerous times.
Such an attempt is also present in the more goal directed path as seen in Figure~\ref{img:path_A}.
One important observation is that sometimes the agent seems to ignore the key, even if the key is very close and in the vision field of the agent.
However, we observed that the agent is able to solve the difficult key puzzle (Figure~\ref{img:double_jump}) that requires the precise execution of two consecutive jumps.
While running 150 episodes per theme and per 3 unseen seeds, these are the measured probabilities for the agent to successfully master this double jump key puzzle using the final model: Ancient: 0\%, Industrial: 30\%, Modern: 33\%.
Based on a shallow search, the agent starts to show the ability of solving this puzzle after about 20,000 PPO updates, with a very low chance of success at the beginning.
With learning progress, the success chance rises substantially on the modern and industrial themes, while staying zero for the ancient one. 
Besides the agent's imminent goals, blue time orbs, which extend the agent's time upon collection, are not really of interest to the agent.
In general, a different behavior is not observed in the case of the agent running out of time.

%% file: content/discussion.tex
%%%%%%%%%%%%%%%%%%%%%%%%%%%%%%%%%%%%%%%%
%%%%%%                        Discussion
%%%%%%%%%%%%%%%%%%%%%%%%%%%%%%%%%%%%%%%%
\section{Discussion}

The performance shown in the evaluation hints that while the agent is able to generalize well on novel seeds on the training themes, it fails to generalize on previously unseen themes.
Therefore, the generalization capability of the employed FFCNN is limited.
On the one hand, it cannot cope well with such strong visual variations as challenged by different OT themes.
On the other hand, the utilized approach manages to train a policy that is able to cope with environment shifts during training.
To some degree, concerning unseen seeds given the 3 training themes, the rather challenging key puzzles that require action sequences like the double jump are solved.
However, high variances and extreme outliers were observed.
These observations and peculiarities of the agent's behavior are discussed and put into context using the subsequent sections to further elaborate the outcome of our work.

%%%%%%%%%%%%%%%%%%%%%%%%%%%%%%%%%%%%%%%%
% Insufficient Visual Observations
\subsection{Limits of Learned Representations}

% shaky movement
% agent likes to go through doors - no matter where they lead to
% agent tries to get past locked doors without having a key
% agents walks across the entire space to grab the key by chance
% key might be ignored by the agent, even if he is close by (hypothesize that frame skipping may be reason for that, agent chooses an action which sticks for few frames (like rotate), and misses frame containing the key while doing so 

Multiple results indicate that the agent's representations are not rich enough to perceive relevant properties and structures of its environment.
As the agent's visual observation is limited to the current and the past two image frames, the policy tells the agent to continuously rotate left and right, while moving forward.
As recurrent hidden states, summarizing previous perceptual history, are missing in the FFCNN, the agent only lives in the moment and thus strives to capture as much relevant information as possible. % in the few buffer frames of its input.
Lacking short-term memory, it has to exploit the capacity of the available RGB frames, which is accomplished by its shaky locomotion behavior.
Besides the game state variables, the image frame stack is the only data available to the agent to make policy decisions.
Therefore, the agent can only react to immediate issues that are contained in its momentary input.
For example, whenever a door is spotted, the agent's imminent behavior is to approach this door no matter where that door leads to.
This may cause the agent to run into doors that are disadvantageous to the agent's goals.
With more information from the short-term past, the agent would gain more potential to improve its navigation.

Another related problem is concerned with the agent being unable to pursue the key or to establish a clear key-door connection.
As previously shown, the agent wanders around entire spaces until it picks up the required key by chance.
By retaining information from the past, such inefficient behavior could be mitigated in a way that the agent does not visit places that he has already visited before.
It could also execute goal-oriented behaviors, for instance recalling positions of an already encountered door after gathering a respective key. % and heading straight to that goal.

Furthermore, the agent sometimes seems to ignore keys that are close to him.
Due to the utilized frame skip, it might be possible that the key is not present on any of the visual observations made by the agent. % when it navigates close to the key.
This becomes more problematic as the key rotates continuously.
Especially due to the low resolution of the visual observations, the key could be missed, because its current visible surface is too small.

Mostly, these issues can be traced back to a lack of short-term memory.
One potential measure towards resolving this problem, without introducing recurrent cells, is to increase the number of image frames and adding the skipped frames to the agent's observation.
Though, collecting more frames raises the dimensionality of the visual observations and makes it more difficult for the training process to efficiently make use of it.
%Thus, a more promising approach is to add a recurrent memory cell (e.g. LSTM and GRU) to the model architecture, as was done by other participants of the OT challenge.

%%%%%%%%%%%%%%%%%%%%%%%%%%%%%%%%%%%%%%%%
% Meaningless Game State Variables
\subsection{Impact of Game State Input Variables}

% is the game state vector input important to the model?
% agent tries to get passed locked doors without having a key
% remaining time and time orbs do not affect its behavior

Frequently, the agent tries to get past locked doors without being in possession of a key.
As the agent directly receives the information whether he has a key or not, it should be able to learn the relationship between these components.
A similar problem can be observed for the time management of the agent.
It neither collects time orbs on purpose nor changes its behavior while running out of time.
It could be possible that the two game state variables alone cannot affect the policy noticeably, because these are just two input features that share the same layer with 3136 dimensional outputs delivered by the visual encoder. % of the model.
In the future, it has to be examined if this hypothesis applies and how to cope with it.
One solution might be to project the game state input to as many units as used by the output of the visual encoder, while sharing the weights among all of those replicated units to avoid parameter explosion.   
Solving this issue may lead to a more robust and better performing policy that is able to consistently exploit the key-door and time bonus relationships.

%%%%%%%%%%%%%%%%%%%%%%%%%%%%%%%%%%%%%%%%
\subsection{Visual Encoder Complexity}

% (visual encoder) model is not complex enough to capture the used skins?
% why is ancient so bad?
    % modern and industrial have a higher contrast and might be biased due to that
% agent is not able to solve the double jump on the ancient theme
% 30% chance on the other two training themes
% should the visual observations be normalized?

Another question emerges concerning the complexity of the visual encoder.
As seen previously, the double jump key puzzle is only solvable for the industrial and modern theme.
The chance of success for the ancient theme is zero.
Further, the agent's performance on this theme is in general inferior to the other ones as illustrated by Figure \ref{img:mean_floor}.
One reason might be that it is more difficult to distinguish the key from ancient themed floor components visually, as in this theme, everything is dyed in brownish hues which impairs object detection.

However, it seems that the difference in performance between the themes becomes expressed very early on in the training, where key puzzles are not introduced yet.
As we performed 3 distinct training runs, we can observe the same scheme each time.
Performance improvements on the modern theme kick in early and rapidly, followed by the industrial and ancient one.
This suggests that this distinct performance is due to differences in some generic visual properties of the training themes.
This visual discrepancy results in a dominating theme that is generally solved better than the others.

%A less likely case could point towards the sampled towers during training.
%If the agent faced  the other themes' double jump puzzles more frequently, it possibly experienced a lack of examples of the mentioned task.
%This theory can be discarded, because the three conducted training runs came to a very similar outcome concerning the inferiorness of the ancient theme, while the modern theme is always the best one.

Now, it can be argued that the model is more likely to extract features from the dominant themes, because those might be more distinguishable, e.g in terms of a stronger contrast.
A more general take on this would be to assume a lack of capacity that impairs the visual encoder to deal with so many distinct themes simultaneously during training.
%It has to be further investigated whether an enlarged or improved visual encoder (e.g, using a deeper ResNet ~\cite{He2016, Espeholt2018} backbone architecture) is able to capture relevant features of more than two skins.
%Another possibility could be to apply a normalization to the image frames that better suits their distribution or data augmentation to make agent more robust towards visual alterations.
%This might help the FFCNN to better accomodate different visual input spaces. This would however require non-trivial knowledge across various domains if training on a substantial number of different theme. 

%%%%%%%%%%%%%%%%%%%%%%%%%%%%%%%%%%%%%%%%
\subsection{Annealing Schedule for Training Parameters}

% optimizing training parameters
% annealing training parameters (learning rate, clip range, entropy bonus coefficient)
        % annealing schedule has to be matched to time span of the learning experience (e.g, so that entropy stays in the loss until agent learns most of the floors ?)
        % does the policy become to deterministic (low entropy)?
        % a lower bound threshold is not used, but would be advisable
        % how to adapt these parameters to the introduction of new tasks?
% gets stuck on corner pieces (extreme outlier)

A problem which was not discussed yet is the agent getting stuck in certain situations by repeating the same action endlessly.
As the entropy bonus coefficient linearly declines over time without a lower bound threshold, the policy may become too deterministic.
If the agent gets stuck at a corner of an exit door, the probability of the action ``moving forward'' is extremely high compared to the other ones.
Thus, the agent can barely escape this situation, which is a possible explanation for the observed negative outliers.

The approach of using annealing training parameters can be questioned in general.
Its intention is to optimize the model stronger in the beginning and fine-tune it towards the end using a lower bound threshold.
This bound was not implemented in this approach and therefore the learning rate, the clip range, and the entropy bonus coefficient equal 0 at the end of the training.
As new subtasks are continually introduced in the environment dependent on reaching higher floors, the agent is required to explore new ways to act, based on how its performance is changing due to environmental shifts.
Making training parameters subject to adaptation from data would further increase the agent's ability to deal with changing tasks.

%% file: content/conclusion.tex
%%%%%%%%%%%%%%%%%%%%%%%%%%%%%%%%%%%%%%%%
%%%%%%                        Conclusion
%%%%%%%%%%%%%%%%%%%%%%%%%%%%%%%%%%%%%%%%
\section{Conclusion and Future Work}

% Conclusion : 
% simple FFCNN networks + PPO agent is able to solve and generalize across seeds, also copes with level shift. However clear limitations, solving tasks with learning pure stimulus-response schemes (door-run through, key-take, etc). Visual encoder does not generalize over different visual themes.

While our original approach performed competitively in the OT challenge, this paper shows that the underlying - rather simple - FFCNN can solve novel seeds on three visual themes, which were faced by the agent during training.
However, the trained model has clear limitations concerning its weak performance across the two left-out visual themes.
Therefore, the visual encoder of the model does not generalize well to unseen visual themes.
By analyzing the agent's behavior on the training themes, it becomes apparent that the agent solves its task in stimulus-response schemes.
This is due to the limited observation space of the agent, where the agent operates on the current and the past two image frames.
By stacking more frames or adding a memory cell to the FFCNN a more proficient agent behavior can be expected.

To improve the generalization capability of the visual encoder, its capacity could be increased by more sophisticated network architectures containing components such as residual blocks ~\cite{He2016} and attention mechanisms~\cite{Vaswani2017}.
It is also advisable to put further effort into understanding the agent's policy.
Using visualization techniques like layer-wise relevance propagation ~\cite{Montavon2019}, it could be possible to derive more insights from saliency maps, which show what inputs are of relevance to the agent ~\cite{Greydanus2018}.
Another concern for further research are adaptive training parameters, like the learning rate. 
This is especially challenging for environments like OT, that encompass multiple subtasks introduced while progressing in the environment.

One drawback to cope with is the slow simulation speed of OT, because it constraints rapid experimenting.
Besides optimizing the environment itself, the developed training process could be made more efficient by augmenting the collected training data.
A further option to accelerate learning experiments is to use distributed training to run DRL algorithm on multiple machines ~\cite{Espeholt2018, Espeholt2020}.
Another approach would be to introduce an adaptive environment that allows the agent to collect the data that is most useful at the current point of training.
For example, if the agent struggles to learn the double jump key puzzle, the puzzle could be made more likely to be visited by the agent, which in general may lead towards exploring different techniques for adaptive sampling. 

% Overall, once more insights are drawn from the aforementioned affairs, the difficult sokoban puzzle, that has only been solved using human demonstrations, can be approached.

%Overall, following the envisioned research directions may result in potent learning algorithms that are also able to cope with more challenging subtasks from scratch like the difficult sokoban puzzle that so far has only been solved using human demonstrations.

Overall, following the envisioned research directions may result in potent learning algorithms that are also able to cope with more challenging subtasks from scratch, like the difficult sokoban puzzle, that has only been solved using human demonstrations so far.

% Future work
    %%%%%%%%%%%%%% LRP and related techniques for visualizing agent choices (e.g as listed here, Visualizing and Understanding Atari Agents, https://arxiv.org/abs/1711.00138)
    % Sokoban puzzle
    %%%%%%%%%%%%%% RNN
    %%%%%%%%%%%%%% Visual encoder with more capacity (e.g ResNet based, or self-attention modules to enable better selective processing depending on env...)
    %%%%%%%%%%%%%%%% Adaptive environment
    %%%%%%%%%%%%%%%% Distributed RL
    %%%%%%%%%%%%%%% Adaptive learning rate (in general, hyperparams like loss weighting for entropy term etc)
    %%%%%%%%%%%%%%%% Data augmentation
    % SAC algorithm? (too specific, maybe not really worth mentioning explicitly)
    %%%%%%%%%%%%%%%% analyze the performance on the evaluation themes
    % ablation study (if environment would be fast enough)
    % entropy bonus suitable exploration method?
    % agent solves double jump after 20k PPO updates (1% chance)

%% file: main.bbl
\begin{thebibliography}{10}
\providecommand{\url}[1]{#1}
\csname url@samestyle\endcsname
\providecommand{\newblock}{\relax}
\providecommand{\bibinfo}[2]{#2}
\providecommand{\BIBentrySTDinterwordspacing}{\spaceskip=0pt\relax}
\providecommand{\BIBentryALTinterwordstretchfactor}{4}
\providecommand{\BIBentryALTinterwordspacing}{\spaceskip=\fontdimen2\font plus
\BIBentryALTinterwordstretchfactor\fontdimen3\font minus
  \fontdimen4\font\relax}
\providecommand{\BIBforeignlanguage}[2]{{%
\expandafter\ifx\csname l@#1\endcsname\relax
\typeout{** WARNING: IEEEtran.bst: No hyphenation pattern has been}%
\typeout{** loaded for the language `#1'. Using the pattern for}%
\typeout{** the default language instead.}%
\else
\language=\csname l@#1\endcsname
\fi
#2}}
\providecommand{\BIBdecl}{\relax}
\BIBdecl

\bibitem{Mnih2015}
V.~Mnih, K.~Kavukcuoglu, D.~Silver, A.~A. Rusu, J.~Veness, M.~G. Bellemare,
  A.~Graves, M.~Riedmiller, A.~K. Fidjeland, G.~Ostrovski, S.~Petersen,
  C.~Beattie, A.~Sadik, I.~Antonoglou, H.~King, D.~Kumaran, D.~Wierstra,
  S.~Legg, and D.~Hassabis, ``Human-level control through deep reinforcement
  learning.'' \emph{Nature}, vol. 518, pp. 529--533, Feb 2015.

\bibitem{Wydmuch2019}
M.~Wydmuch, M.~Kempka, and W.~Jaskowski, ``Vizdoom competitions: Playing doom
  from pixels,'' \emph{{IEEE} Trans. Games}, vol.~11, no.~3, pp. 248--259,
  2019.

\bibitem{Jaderberg2019}
M.~Jaderberg, W.~M. Czarnecki, I.~Dunning, L.~Marris, G.~Lever, A.~G.
  Castañeda, C.~Beattie, N.~C. Rabinowitz, A.~S. Morcos, A.~Ruderman,
  N.~Sonnerat, T.~Green, L.~Deason, J.~Z. Leibo, D.~Silver, D.~Hassabis,
  K.~Kavukcuoglu, and T.~Graepel, ``Human-level performance in 3d multiplayer
  games with population-based reinforcement learning.'' \emph{Science (New
  York, N.Y.)}, vol. 364, pp. 859--865, May 2019.

\bibitem{OpenAI2019}
OpenAI, C.~Berner, G.~Brockman, B.~Chan, V.~Cheung, P.~Debiak, C.~Dennison,
  D.~Farhi, Q.~Fischer, S.~Hashme, C.~Hesse, R.~J{\'o}zefowicz, S.~Gray,
  C.~Olsson, J.~Pachocki, M.~Petrov, H.~P. de~Oliveira~Pinto, J.~Raiman,
  T.~Salimans, J.~Schlatter, J.~Schneider, S.~Sidor, I.~Sutskever, J.~Tang,
  F.~Wolski, and S.~Zhang, ``Dota 2 with large scale deep reinforcement
  learning,'' \emph{arXiv:1912.06680}, 2019.

\bibitem{Baker2019}
B.~Baker, I.~Kanitscheider, T.~Markov, Y.~Wu, G.~Powell, B.~McGrew, and
  I.~Mordatch, ``Emergent tool use from multi-agent autocurricula,''
  \emph{arXiv:1909.07528}, 2019.

\bibitem{Vinyals2019}
O.~Vinyals, I.~Babuschkin, W.~M. Czarnecki, M.~Mathieu, A.~Dudzik, J.~Chung,
  D.~H. Choi, R.~Powell, T.~Ewalds, P.~Georgiev, J.~Oh, D.~Horgan, M.~Kroiss,
  I.~Danihelka, A.~Huang, L.~Sifre, T.~Cai, J.~P. Agapiou, M.~Jaderberg, A.~S.
  Vezhnevets, R.~Leblond, T.~Pohlen, V.~Dalibard, D.~Budden, Y.~Sulsky,
  J.~Molloy, T.~L. Paine, C.~Gulcehre, Z.~Wang, T.~Pfaff, Y.~Wu, R.~Ring,
  D.~Yogatama, D.~Wünsch, K.~McKinney, O.~Smith, T.~Schaul, T.~Lillicrap,
  K.~Kavukcuoglu, D.~Hassabis, C.~Apps, and D.~Silver, ``Grandmaster level in
  starcraft ii using multi-agent reinforcement learning.'' \emph{Nature}, Oct.
  2019.

\bibitem{Schulman2017}
J.~Schulman, F.~Wolski, P.~Dhariwal, A.~Radford, and O.~Klimov, ``Proximal
  policy optimization algorithms,'' \emph{arXiv:1707.06347}, 2017.

\bibitem{Juliani2019}
A.~Juliani, A.~Khalifa, V.~Berges, J.~Harper, E.~Teng, H.~Henry, A.~Crespi,
  J.~Togelius, and D.~Lange, ``Obstacle tower: {A} generalization challenge in
  vision, control, and planning,'' in \emph{Proceedings of the 28th
  International Joint Conference on Artificial Intelligence, {IJCAI} 2019},
  2019, pp. 2684--2691.

\bibitem{Juliani2019_blog}
A.~Juliani and J.~Shih, ``Announcing the obstacle tower challenge winners and
  open source release,'' 2019, available at
  \url{https://blogs.unity3d.com/2019/08/07/announcing-the-obstacle-tower-challenge-winners-and-open-source-release/}
  retrieved March 20, 2020.

\bibitem{Nichol2019}
A.~Nichol, ``Competing in the obstacle tower challenge,'' 2019, available at
  \url{https://blog.aqnichol.com/2019/07/24/competing-in-the-obstacle-tower-challenge/}
  retrieved March 24, 2020.

\bibitem{Booth2019}
J.~Booth, ``{PPO} dash: Improving generalization in deep reinforcement
  learning,'' \emph{arXiv:1907.06704}, 2019.

\bibitem{Bishop2006}
C.~M. Bishop, \emph{Pattern Recognition and Machine Learning (Information
  Science and Statistics)}.\hskip 1em plus 0.5em minus 0.4em\relax Secaucus,
  NJ, USA: Springer-Verlag New York, Inc., 2006.

\bibitem{zhang2018}
C.~Zhang, O.~Vinyals, R.~Munos, and S.~Bengio, ``A study on overfitting in deep
  reinforcement learning,'' \emph{arXiv:1907.06704}, 2018.

\bibitem{perez2019general}
D.~Perez-Liebana, J.~Liu, A.~Khalifa, R.~D. Gaina, J.~Togelius, and S.~M.
  Lucas, ``General video game ai: A multitrack framework for evaluating agents,
  games, and content generation algorithms,'' \emph{IEEE Transactions on
  Games}, vol.~11, no.~3, pp. 195--214, 2019.

\bibitem{pcgbook}
N.~Shaker, J.~Togelius, and M.~J. Nelson, \emph{Procedural Content Generation
  in Games}.\hskip 1em plus 0.5em minus 0.4em\relax Springer, 2016.

\bibitem{Justesen2018}
N.~Justesen, R.~R. Torrado, P.~Bontrager, A.~Khalifa, J.~Togelius, and S.~Risi,
  ``Illuminating generalization in deep reinforcement learning through
  procedural level generation,'' \emph{arXiv:1806.10729}, 2018.

\bibitem{Cobbe2019a}
K.~Cobbe, O.~Klimov, C.~Hesse, T.~Kim, and J.~Schulman, ``Quantifying
  generalization in reinforcement learning,'' in \emph{Proceedings of the 36th
  International Conference on Machine Learning, {ICML} 2019, 9-15 June 2019,
  Long Beach, California, {USA}}, ser. Proceedings of Machine Learning
  Research, vol.~97.\hskip 1em plus 0.5em minus 0.4em\relax {PMLR}, 2019, pp.
  1282--1289.

\bibitem{Cobbe2019}
K.~Cobbe, C.~Hesse, J.~Hilton, and J.~Schulman, ``Leveraging procedural
  generation to benchmark reinforcement learning,'' \emph{arXiv:1912.01588}.

\bibitem{FortunatoNIPS2019}
M.~Fortunato, M.~Tan, R.~Faulkner, S.~Hansen, A.~Puigdom\`{e}nech~Badia,
  G.~Buttimore, C.~Deck, J.~Z. Leibo, and C.~Blundell, ``Generalization of
  reinforcement learners with working and episodic memory,'' in \emph{Advances
  in Neural Information Processing Systems 32}.\hskip 1em plus 0.5em minus
  0.4em\relax Curran Associates, Inc., 2019, pp. 12\,469--12\,478.

\bibitem{HofmannNIPS2019}
M.~Igl, K.~Ciosek, Y.~Li, S.~Tschiatschek, C.~Zhang, S.~Devlin, and K.~Hofmann,
  ``Generalization in reinforcement learning with selective noise injection and
  information bottleneck,'' in \emph{Advances in Neural Information Processing
  Systems 32}.\hskip 1em plus 0.5em minus 0.4em\relax Curran Associates, Inc.,
  2019, pp. 13\,978--13\,990.

\bibitem{lee2019network}
K.~Lee, K.~Lee, J.~Shin, and H.~Lee, ``Network randomization: A simple
  technique for generalization in deep reinforcement learning,''
  \emph{arXiv:1910.05396}, 2019.

\bibitem{James2019}
S.~James, P.~Wohlhart, M.~Kalakrishnan, D.~Kalashnikov, A.~Irpan, J.~Ibarz,
  S.~Levine, R.~Hadsell, and K.~Bousmalis, ``Sim-to-real via sim-to-sim:
  Data-efficient robotic grasping via randomized-to-canonical adaptation
  networks,'' in \emph{Proceedings of the IEEE Conference on Computer Vision
  and Pattern Recognition}, 2019, pp. 12\,627--12\,637.

\bibitem{gym_minigrid}
M.~Chevalier-Boisvert, L.~Willems, and S.~Pal, ``Minimalistic gridworld
  environment for openai gym,'' \url{https://github.com/maximecb/gym-minigrid},
  2018.

\bibitem{Bellemare2013}
M.~G. Bellemare, Y.~Naddaf, J.~Veness, and M.~Bowling, ``The arcade learning
  environment: An evaluation platform for general agents,'' \emph{J. Artif.
  Intell. Res.}, vol.~47, pp. 253--279, 2013.

\bibitem{Sutton2018}
R.~S. Sutton and A.~G. Barto, \emph{Reinforcement learning: An
  introduction}.\hskip 1em plus 0.5em minus 0.4em\relax MIT press, 2018.

\bibitem{Mnih2016}
V.~Mnih, A.~P. Badia, M.~Mirza, A.~Graves, T.~Lillicrap, T.~Harley, D.~Silver,
  and K.~Kavukcuoglu, ``Asynchronous methods for deep reinforcement learning,''
  in \emph{International conference on machine learning}, 2016, pp. 1928--1937.

\bibitem{Tavakoli2017}
A.~Tavakoli, F.~Pardo, and P.~Kormushev, ``Action branching architectures for
  deep reinforcement learning,'' in \emph{Proceedings of the 32nd {AAAI}
  Conference on Artificial Intelligence}.\hskip 1em plus 0.5em minus
  0.4em\relax {AAAI} Press, 2018, pp. 4131--4138.

\bibitem{He2016}
K.~{He}, X.~{Zhang}, S.~{Ren}, and J.~{Sun}, ``Deep residual learning for image
  recognition,'' in \emph{Proc. IEEE Conf. Computer Vision and Pattern
  Recognition (CVPR)}, Jun. 2016, pp. 770--778.

\bibitem{Vaswani2017}
A.~Vaswani, N.~Shazeer, N.~Parmar, J.~Uszkoreit, L.~Jones, A.~N. Gomez,
  L.~Kaiser, and I.~Polosukhin, ``Attention is all you need,'' in
  \emph{Advances in Neural Information Processing Systems 30: Annual Conference
  on Neural Information Processing Systems 2017, 4-9 December 2017, Long Beach,
  CA, {USA}}, 2017, pp. 5998--6008.

\bibitem{Montavon2019}
G.~Montavon, A.~Binder, S.~Lapuschkin, W.~Samek, and K.~M{\"{u}}ller,
  ``Layer-wise relevance propagation: An overview,'' in \emph{Explainable {AI:}
  Interpreting, Explaining and Visualizing Deep Learning}, ser. Lecture Notes
  in Computer Science.\hskip 1em plus 0.5em minus 0.4em\relax Springer, 2019,
  vol. 11700, pp. 193--209.

\bibitem{Greydanus2018}
S.~Greydanus, A.~Koul, J.~Dodge, and A.~Fern, ``Visualizing and understanding
  atari agents,'' in \emph{International Conference on Machine Learning}, 2018,
  pp. 1792--1801.

\bibitem{Espeholt2018}
L.~Espeholt, H.~Soyer, R.~Munos, K.~Simonyan, V.~Mnih, T.~Ward, Y.~Doron,
  V.~Firoiu, T.~Harley, I.~Dunning, S.~Legg, and K.~Kavukcuoglu, ``{IMPALA:}
  scalable distributed deep-rl with importance weighted actor-learner
  architectures,'' in \emph{Proceedings of the 35th International Conference on
  Machine Learning}, ser. Proceedings of Machine Learning Research,
  vol.~80.\hskip 1em plus 0.5em minus 0.4em\relax {PMLR}, 2018, pp. 1406--1415.

\bibitem{Espeholt2020}
L.~Espeholt, R.~Marinier, P.~Stanczyk, K.~Wang, and M.~Michalski, ``Seed rl:
  Scalable and efficient deep-rl with accelerated central inference,'' in
  \emph{International Conference on Learning Representations}, 2020.

\end{thebibliography}
